\documentclass{article}
\usepackage{graphicx} 
\usepackage{url}

\title{\bf Amplifying Limitations, Harms and Risks of Large Language Models}
\author{Michael O'Neill \& Mark Connor\\
UCD Natural Computing Research \& Applications Group\\
School of Business\\ University College Dublin\\
m.oneill@ucd.ie, mark.connor@ucd.ie}
\date{}

\begin{document}
\maketitle

\begin{abstract}
    We present this article as a small gesture in an attempt to counter what appears to be exponentially growing hype around Artificial Intelligence (AI) and its capabilities, and the distraction provided by the associated talk of science-fiction scenarios that might arise if AI should become sentient and super-intelligent. It may also help those outside of the field to become more informed about some of the limitations of AI technology. In the current context of popular discourse AI defaults to mean foundation and large language models (LLMs) such as those used to create ChatGPT. This in itself is a misrepresentation of the diversity, depth and volume of research, researchers, and technology that truly represents the field of AI. AI being a field of research that has existed in software artefacts since at least the 1950's. We set out to highlight a number of limitations of LLMs, and in so doing highlight that harms have already arisen and will continue to arise due to these limitations. Along the way we also highlight some of the associated risks for individuals and organisations in using this technology.
\end{abstract}

\section{Introduction}
With at least one author of this article being a researcher in Artificial Intelligence (AI)~\cite{mitchell:aguideforthinkinghumans} for over 25 years, and all having undertaken PhDs in an evolutionary form of generative AI, based on formal generative grammars~\cite{mckayetal:2010}, the recent advances made in the development of large language models (LLMs) and their associated tools are perhaps the most impressive and surprising results we have witnessed during our careers. These advances and the provision of access to the general public to use the arising technology has resulted in a lot of media interest and hype, which is tending to over-emphasise the possibility of futuristic, science fiction-like scenarios occurring. This is often at the expense of masking the fundamental limitations of the existing technology, the resulting harms which it can and has caused, and the future long-term risks associated with continuing to blindly adopt the technology. On a more positive note, it is nice to see the field of AI emerge from its relative obscurity into the mainstream where it seems everyone now has an interest and opinion. 

Briefly, for the benefit of readers without a background in AI, and at the risk of introducing many many more confusing (three letter) acronyms, let's say a few words about the relationship between LLMs and AI. Large language (LLM) and foundation model technology in their current form are based on deep learning, which belongs to a sub-field of AI known as machine learning (ML)~\cite{mitchell:1997}. Deep learning is effectively a re-branding and development of what were traditionally called multi-layer perceptrons (MLP)~\cite{rummelhartmcclelland}, a form of artificial neural network (ANN). ANNs and MLPs are inspired by the biological brain, and have developed out from a model of a single brain cell (a neuron) called a perceptron~\cite{mcculloghpitts}. There are in fact many AI technologies that have been inspired by nature, with researchers also attempting to distil the computational essence underpinning, for example, the biological process of evolution, the social behaviour of insects, birds and fish, the immune system, the grammar underpining language, physics (classical and quantum) and even chemistry~\cite{brabazononeillmcgarraghy}. 


One can visualise a deep learning model as being a very large network of computational units (analagous to "brain cells") comprised of many connected layers, with each layer itself being comprised of many computational units. The machine learning process adjusts the importance (weights) of the connections between the different computational units in response to network outputs generated by the data upon which the network is trained. Researchers in machine learning have and continue to explore, for example, how to implement the computational units, how to learn (or adjust the weights on the network connections), and how to design the overall architecture of the network. Coupled to massive amounts of training data, and to increasingly fast computer number crunching power upon which these deep networks can run, a pivotal transition moment in the development of the deep learning technology that underpins the LLM (GPT) behind ChatGPT was the invention of the transformer network architecture~\cite{vaswanietal:2017}. For the interested reader in finding out more about the field of AI, as an accessible and recent introduction we would recommend Mitchell's book~\cite{mitchell:aguideforthinkinghumans}. In terms of gaining some perspectives on the history of the field, a nice collection of articles can be found in~\cite{lungarella:etal:2007}.

In the remainder of this article we attempt to amplify the limitations, harms and risks of LLMs, and foundation models more generally, in an attempt to re-balance the current discourse, and ideally help to inform and provide those outside of the machine learning and AI community (and perhaps some within) with a more holistic perspective and awareness that this technology should not be blindly accepted and adopted. Before discussing the limitations of LLM technology let's first say a few words about ChatGPT and the LLM that underpins it.


\section{A brief focus on the Large Language Model of ChatGPT}
We think it is fair to say that the coherence of the text generated by the likes of ChatGPT has surprised many researchers. Particularly the apparent jump in performance from November 2022 (GPT-3.5) to March 2023 (GPT-4). Unfortunately, we do not have all the necessary information required to properly analyse how ChatGPT is working under the hood. Since Microsoft invested a billion dollars in OpenAI in 2019, and then ten billion dollars more in January 2023, they have been less transparent, but GPT-4 does appear to have raised its game over and above previous LLM-based tools. Why? 

Working through the GPT-4 Technical Report~\cite{GPT-4-TR} and various media reports, some factors to explain performance gains include:
\begin{itemize}
    \item the model may be bigger than GPT-3's 175 billion parameters~\cite{GPT-3-2020}, with numbers speculated to be somewhere around the one trillion region. An alternative hypothesis is that GPT-4 is a mixture of models (8 models each containing 220 billion parameters) not one monolithic model. As an aside, it is reported to have cost more than US\$100 Million to train it.
    \item GPT-4 can be given a larger context of text to set the scene. The pay-walled GPT-4 has 32K tokens vs 8K for standard GPT-4, while for the GPT-3 generation it started with a context of 2k, which then grew to 4k tokens. 
    \item GPT-4 may have been trained using significantly more tokens, a trend that has also been seen in Google's PaLM2 model which was reportedly trained on 3.6 trillion tokens, 2.82 trillion more than its predecessor PaLM.  
    \item sitting on top of the GPT-4 large language model is a stack of added layers of rules and engineering, which implement guard rails to try and filter the generated output to avoid some of the factually incorrect, harmful, and biased text that can be generated. 
    \item there is also an additional layer of learning sitting on top of the LLM, called reinforcement learning through human feedback (RLHF). The purpose of this layer is to align the model's output with human preferences. 
\end{itemize}

On that last point, it is worth highlighting that our (the public) interactions with ChatGPT can be used to refine any current or future OpenAI models based on the feedback we provide. Currently, this feedback is discretely collected in the form of a thumbs up or down vote next to all ChatGPT responses. Users in that case are effectively working (for free) for OpenAI, or if they are using the pay-walled version, users are paying to work, enabling OpenAI and Microsoft to create a better product by way of refinement based on the user's feedback.

This also raises a potentially serious security and privacy risk. There exists the possibility that users are sharing sensitive and/or personal identity information or perhaps even sensitive government information \cite{RTE-05152023} with OpenAI. For example, what if ChatGPT is behind the customer service chatbot on your utilities provider, insurance broker, or bank web interface or app? You might think that you're engaging with a trusted service provider, and as such, you are comfortable sharing dates of births, payment card details, your address, etc. If ChatGPT, or whatever LLM, is being used by your service provider as a "software-as-a-service" (SaaS) by the developers of the LLM, there is the possibility that all your details have automatically been passed to that third party! From an EU GDPR perspective, the LLM SaaS provider is now a data processor. Have you given them permission to process your data? What are they doing with your data? Are they storing it? Do the developers of the LLM see examples of the prompts coming in to the LLM in order to improve its guardrails etc? Can those developers be trusted with your data? Is your data being incorporated into the RLHF layer of the model? With an appropriately designed prompt could your data then be retrieved and regurgitated back to another party, perhaps a bad actor, who might then use your data for fraud?

Clearly, without having delved too deep into the technology behind the likes of ChatGPT we start to encounter some uncomfortable questions and risks. Given the significant hype that currently exists around ChatGPT and LLMs generally, in this article we set out to provide a more sober perspective on the technology, by highlighting a number of issues with LLMs. In so doing, we are not claiming novelty in our perspective. Rather in many cases we are simply amplifying issues that have already been raised by other researchers drawn from our field's wider core set of disciplines (e.g., Bender et al~\cite{bender_koller_2020,benderetal_2021}, Melanie Mitchell~\cite{pnas2023,melaniemitchell:blog}, Abeba Birhane~\cite{birhane:phdthesis}, Judea Pearl~\cite{pearl}, Rodney Brooks~\cite{brooks:blog}, etc.,). This is also by no means intended to be an attack on ChatGPT, LLM and foundation model technology. It would be foolish to dismiss the potential utility of this technology, particularly as a productivity tool when used by a human, where the human is filtering, correcting, and taking responsibility for the output. But given the level of hype that currently exists, coupled with the wide accessibility of the technology, there is a real and growing danger that unsuspecting users will be caught off guard (e.g., recent reports on a US Lawyer using ChatGPT for legal research~\cite{BBC-65735769}), that harm will continue to arise, and that risks will not be recognised without a number of researchers joining in to counter the hype by amplifying what might be considered a more grounded scientific perspective that identifies the limitations of this technology (e.g., for discussions on their risk of use, see Birhane et al~\cite{BirhaneNatRevPhys:2023} on use in Science, Morley \& Floridi~\cite{MorleyFloridi:2023} on their use in Medicine, and Gros et al~\cite{Grosetal:2023} on their use in Software Engineering). 

At this point it is also important to mention that ChatGPT is but one example of a LLM-based technology. There exists a rapidly expanding ecosystem of large language and foundation models that can generate text, code, images, video and audio~(e.g.,~\cite{bert,lambda,dall-e-2,stablediffusion,github-copilot,make-a-video,musiclm}). Also, recognising that in terms of foundation models such as LLM's that the "genie is out of the bottle". That is, various models are in the public domain, and are becoming increasingly accessible with variants emerging that are smaller and capable of running on personal devices. As such these models are not going to disappear if the big corporations switch off access to them. As many of the commercial organisations deploying LLM and foundation model technology are more focused on effective marketing hype and sales, it unfortunately falls upon users to educate themselves about this technology. To this end, in the remainder of this article we outline some of the main limitations that we and others have observed about LLMs and foundation models. 

\section{What are the Limitations, Harms and Risks of Large Language Models?}
A number of fundamental limitations can be associated with LLM-based technology and more broadly foundation models, which in turn can give rise to harms and risks. These include but are not limited to:
\begin{enumerate}
\item Meaning arises from expression with intent
\item Generative stochasticity
\item Anthropomorphism
\item The training data contains biases, nonsense and harmful content
\item Security and Privacy
\item Correlation is not causation
\item All models are wrong, some are useful
\item Leaky data makes it hard to evaluate true performance
\item Emergent abilities are misleading
\item Rights-laundering
\item Sustainability
\item Regulatory non-compliance
\item Transparency
\item Equality, Diversity \& Inclusion
\end{enumerate}

\noindent Let's briefly touch upon each of the limitations of LLM technology listed above.

\subsection{Meaning arises from expression with intent}

Put simply, LLMs do not understand the prompts given to them, and do not understand the text they generate.
From a Linguistics perspective, researchers like Bender \& Koller~\cite{bender_koller_2020}, highlight that the data upon which LLMs are trained does not contain intent that is present in human communication. This lack of intent means that only very weak semantics, or meaning, can be captured by a LLM trained on purely expression (structural) data. To think of this another way, when humans communicate they do so in such a way that there is (hopefully) a shared understanding of the intent and meaning of the communication. This arises in part due to joint attention, where those engaged in the communication are focused on the same subject. They also have (shared) life experience and an associated world model (including practical knowledge of concepts, such as gravity, that arise due to the physics our world exists in, as well as other cultural and social signals), which would include a model of self (knowledge of our own sensorimotor space), a theory of mind, and those also are likely sensing other signals outside of the raw text of the message such as body language, intonation, a shared physical context etc. These are simply not present in a ChatGPT or similar technology.  
Human understanding and meaning appears to arise from a grounding of language in concepts through joint attention, embodiment and associated world model, and perhaps multimodality may also be required (see literature on embodied cognition e.g.,~\cite{embodiedcognition:stanford,tamari}). 

In other words, on their own LLMs at present don't understand, and they are not sentient. For LLMs to gain understanding, either more sophisticated LLMs will need to be developed that successfully capture intent, and/or additional "layers" of capability will be required that will somehow implement understanding. How that might be achieved is very much an open research problem. For the current generation of LLM-based technology, the human user is the language model layer that is implementing all the understanding by conferring meaning onto the prompts and generated text.

\subsection{Generative stochasticity}

The nature of the process used to generate the text outputted by a large language model is that it includes a degree of stochasticity (i.e., randomness). Bender et al~\cite{benderetal_2021} have referred to LLMs as Stochastic Parrots. We can consider a LLM such as ChatGPT to be a complex ``auto-complete" tool. That is, given a list of words (the prompt) provided by the user, the LLM passes that prompt through its model to generate the next word that might follow the prompt, and then the next word following that, and the next word following that word, and so on. Note that we simplify this description to ease discussion, so that, for example, we draw an equivalence between tokens (the lowest level primitives that LLMs process and generate) and words. When selecting the next word to output, there may be a number of words available to the LLM that might be equal or similar in likelihood to occur. This is where the randomness comes into play. One of these words is chosen at random. If you have ever used ChatGPT, this explains why if you give the same prompt over and over again, each time the response will be different. This is generative stochasticity at work. Coupled with the inability to understand meaning, generative stochasticity also plays its part in the generation of nonsensical output. LLM output is strongly linked to the training data used to create the LLM. That is, an LLM will never make a next-word prediction that it has not seen in the training data, and conversely, it favours those words which are more strongly represented in the training data. A truly intelligent and creative agent/system should be capable of selecting a word based on first-order logical and deductive reasoning thus reducing the generation of nonsensical output to nearly zero, as opposed to the current mechanism LLMs employ which is based on probabilistic stochastic selection, adding weight to our categorisation of LLMs as complex ``auto-complete" tools.

\subsection{Anthropomorphism}

We humans are flawed, we have many cognitive biases~\cite{kahneman}. Perhaps AI might be useful in the future to assist us in counteracting our cognitive biases. When engaging with a chatbot (underpinned by an LLM or even some other language model) through an interface where we enter a prompt, and the bot replies by seemingly typing out letter by letter, word by word a response, perhaps even pausing before starting to respond. The delay in response may be deliberately engineered by the chatbot designers, or it could arise as a result of the large amount of processing that might be required to generate a response due to the size of the LLM. The chatbot interface may even provide a human-like name to the chatbot agent. In such an environment, even if we are told the chatbot is a computer program and not a human, we can still succumb to an "Eliza effect"~\cite{elizaeffect}. Our learned ability to process and find meaning in conversational text means that we slip, and often forget that we are not dealing with a human, and very quickly we anthropomorphise the bot. We may attribute personality traits to it. We may attribute an intended meaning to the generated text. Anthropomorphism of this nature is not trivial and can have serious consensuses, indeed previous research has highlighted people's tendency to ``overtrust" robotic systems with potentially disastrous outcomes \cite{Robinette2016}.   

Arguably the designers of tools such as ChatGPT should provide interfaces that make it much clearer (and perhaps in a repetitive, ongoing basis) to the human user that this is a bot. They should also be careful with the language they use to describe their models, systems and services, avoiding terms that might be associated with human abilities. However, the tools are designed to be appealing to the human user by being human-like. It also does not help that anthropomorphic terms are used to describe nonsensical output, such as describing this as the model sometimes "hallucinates", rather than just clearly stating that it can make stuff up, generating text that is "factually incorrect" or simply "nonsense".    

\subsection{The training data contains biases, nonsense and harmful content}

Due to the data upon which many of the LLMs that currently exist are trained, the output generated by LLMs may contain harmful, biased, intolerant, hate and factually incorrect content.
The reality is, that in the case of ChatGPT, we simply do not know what the underlying training data set is that was used to develop the GPT-4 LLM, as OpenAI have not shared this information. We know that GPT-3 was trained on a combination of sources drawn from the wilds of the internet, including Common Crawl data from 2016 to 2019, Webtext, Books and Wikipedia~\cite{GPT-3-2020}. Analysis of Common Crawl data has shown that it contains harmful content, such as hate speech and sexually explicit content (e.g., \cite{Luccioni2021}). We're basically talking about the training data including the content on web pages drawn from across the world wide web. As such it is not surprising if we find mis- and dis-information, stereotypes and bias, falsehoods, hate speech, sexist, and racist content, etc., in the training data.

Given the content of the training data, coupled with the power of these models to faithfully capture the training data, and the stochastic nature of these models, it is always possible that harmful content, bias, mis- and dis-information, and outright fabrications will be output. As Kidd \& Birhane~\cite{kiddbirhane:2023:science} highlight, these models then have the potential to distort human beliefs. And, this is reinforced by the hype that is associated with the technology, in its overstating its capability. And, further driven by our human tendency to anthropomorphise the technology and its ability, as discussed earlier. To quote Kidd \& Birhane \begin{quote}
    "People form stronger, longer-lasting beliefs when they receive information from agents that they judge to be confident and knowledgeable..." \cite{kiddbirhane:2023:science}
\end{quote} 

Attempts to refine the outputs resulting from the LLM training data have been made using techniques such as RLHF. Although effective in part, the RLHF process manifests its own issues regarding biases and content. There exists a subtle trade off between the inconsiderate curating of free speech and the moderation of harmful content. The lines separating these two actions are not always so black and white, they likely depend heavily on ones own beliefs, culture, religion and heritage, etc. The creators of LLMs and technologies derived from them are currently the ones who draw these lines, thus making it likely that these models will inherently reflect their own morals and beliefs, given that the four most prominent models in production today have been developed by companies founded and head quartered in the United States of America. One might postulate that attempts to remove biases, nonsense and harmful content from the training data may also come with certain degrees of diversity and cultural suppression by way of amplifying these companies own beliefs and traditions. 
\subsection{Security \& Privacy}
As discussed in the introduction, there is the possibility of sensitive data being passed to a third party providing the LLM as SaaS, and/or sensitive data being ingested into the model during model training and later massaged out of it through prompt engineering. Recent work by \cite{Huang2022} has highlighted the propensity of Large Pre-Trained Language Models to leak private or sensitive information contained within their training data corpus. These leaks are possible because of the model's ability to "memorize" sequences of tokens at training time \cite{Huang2022, Carlini2020, Carlini2022}. Ironically, Huang et al, also found that the models' inability to associate related pieces of information (e.g. Name, Email Address, etc) made it more difficult for attackers to extract the information. Nonetheless, this fragility poses the real and significant threat of exposing the personal information of individuals contained in an LLM training data set, information that individuals may not have agreed to share for the original or otherwise intended purpose of training LLMs.
\subsection{Correlation is not causation}

The language of statistics is not amenable to capturing causation (e.g., see~\cite{pearl}), and correlations do not imply causation. The current generation of LLMs are effectively statistical models of token/word associations, and as such are models of correlation. These and other foundation models do not capture the Why or causation of association. As such they are not capable of causal reasoning.

\subsection{All models are wrong, some are useful}

All models are simplifications of the reality they attempt to represent. LLMs are not an exception to this. Simplifications can arise in the representative capacity of the deep learning approach (the computation being undertaken at each node in the network, the parameter ranges and algorithm strategies e.g., the type and rate of learning), and as highlighted earlier there are clearly issues with the representative nature of the training data. The training data does not provide perfect coverage of all possible sentences that might be expressed in every human language, in fact, this is impossible given that is an infinite space! Another interesting feature of human language, is that it changes over time, it might be said to evolve through the generations of humans that live the language in an evolving society~\cite{languageevolution}. To be relevant in terms of the latest developments in, for example, science and society, and to embrace the adaptive nature of the language itself, LLMs will need to be periodically re-trained. Given the estimated cost of training ChatGPT being in the order of 100 Million US dollars, and the duration that it takes to run the training (months), this is not a task that will be undertaken lightly or frequently, and as such the relevance of the underlying language model will start to decay from the moment training stops. As to the rate of decay and when it would stop being useful, will likely depend on the context in which it is being used, and also in proportion to the rate of the change language, the rate of generation of scientific knowledge, the rate of generation of artistic and cultural output, and the rate of generation of news, etc.

\subsection{Leaky data makes it hard to evaluate true performance}

We often do not know what comprises the data used to train LLMs. For example, OpenAI have not revealed the training data for GPT-4. This presents a significant challenge to rigorously evaluate the performance of these models in a scientific manner. The traditional way to measure the performance of a (machine learning) model is to test the model on "unseen data". That is data that the model has not been trained on. We refer to this as the test (data) set. This test set performance measurement sets out to determine the level of ability the model might have to generalise beyond what it has been trained on. For example, has the model learned general concepts and principles that can be applied to a wider set of circumstances? But if the test set contains data that is included in the training set, we have what is known as "leaky data", and we cannot be sure if the model is simply regurgitating  what it has already been exposed to, akin to a photographic memory without necessarily understanding the meaning of what it has stored in its memory. Interestingly, the proposed amendments to the EU AI Act, which are currently working their way through the European Parliament, will require those who deploy LLMs to reveal the training data used by the LLM.

\subsection{Emergent Abilities are misleading}
Recently the term "emergence" has started to be strongly associated with the abilities of large language models. This term has many interpretations but in the context of LLMs emergence generally refers to abilities or behaviours that are not present in smaller models but are present in larger ones, such that these abilities or behaviours could not have been predicted by simply extrapolating the performance of smaller models \cite{Wei2022, Schaeffer2023}. The idea that LLMs have surpassed some necessary scale needed to give rise to unpredictable and inherently uncontrollable emergent abilities has fuelled much of the rhetoric and hysteria surrounding their future potential to cause catastrophic harm by acting autonomously and adaptability through an agent of some kind. At the moment this is the stuff of science fiction and distracts from the very real potential for harm that LLMs present. Whilst the claims of LLM emergence continue to prevail little scientific analysis has been conducted to evaluate the type and true extent of emergence in LLMs. Initial work has postulated that emergence in LLMs can be largely explained by the choice of metric a researcher uses to evaluate fixed model outputs \cite{Schaeffer2023}. The author's findings concluded that when using the InstructGPT/GPT3 family of models, the presence of emergent abilities or behaviours could be modulated through the use of different metrics and improved statistical analysis, thus pouring cold water on the theory that the presence of emergence in LLMs is a property of their scale. Other theories relating to the "emergence" of the human-like reasoning and understanding abilities sometimes exhibited by LLMs also support a much more grounded statistical explanation suggesting that LLMs exploit the distributional properties of language in latent space \cite{Jiang2023, Wang2023, Xie2021}. While research is ongoing to explain the emergent-like behaviour of LLMs, all be it hampered by the lack of access to propriety models and data, it is important not to be misled by claims that LLMs do indeed have inherent emergent abilities and by extension could be capable of developing enhanced capabilities in a random uncontrollable fashion or produce outputs which are the result of an ``artificial intelligence" which has emerged during the models development process.    

Another perspective on emergence, arises when a technology such as a LLM becomes part of a larger technology solution or system. What might be the unintended consequences that might emerge, especially in light of the various limitations of the underlying LLM itself? How might these limitations be expressed and ripple through the behaviour of the larger system of which it is part of? 

\subsection{Rights-laundering}

The training data of many foundation and large language models include copyrighted material and material published under licences with varying permissiveness. It is entirely possible, and many examples have been reported, where copyrighted or licensed material is retrieved or reproduced as the output of a LLM. Lawsuits are even being filed as a result~\cite{Samuelson:2023}. 

Who owns the output of a LLM or other foundation model? Is it the person entering the prompt? Is it the people or organisation responsible for the foundation model? Does the foundation model own it? Or is it jointly owned? If the original copyright or licence holder is not protected this effectively amounts to licence laundering or copyright laundering. 

In a disturbing twist, it is possible to use some of these foundation models to generate voice, image, and video versions of individuals (deep fakes). Effectively an individual person can be impersonated through media whether or not they give permission for this. For what purpose might this be adopted, and by whom? How will society at large be capable of distinguishing the virtual from the reality?

This potential for rights-laundering does suggest that regulations are required as soon as possible and that greater responsibility be placed on the developers of these technologies to limit misuse.

If for a moment we are to overlook the potential breaches of law that might arise in the use of this technology. In light of the potential for rights laundering, we should be asking ourselves if it is good ethical practice to automatically assume that we can adopt the output of a LLM as if we own it and have the right to use it. {\bf \em Just because we can use the tool and its output}, we should be asking ourselves, {\bf \em should we use the tool and its output}?

\subsection{Sustainability}

Bender et al~\cite{benderetal_2021} highlight the unsustainable nature of the current generation of LLM technology, given the significantly increasing amount of energy and resources required to train and deploy LLM-technology. They also note the environmental racism that arises when the world's most marginalised communities are the ones most deeply impacted by climate change arising from technology development and use by more privileged communities. We have to ask ourselves are we prepared to accept the environmental and societal cost? Or should we be focused on developing more sustainable AI technology?

\subsection{Regulatory non-compliance}
We mentioned earlier the potential of foundation model technology for rights-laundering in the output it might generate. Thankfully, efforts are already underway to move to regulate AI technology. The EU are one of the early movers in this regard having initially developed ``Ethics guidelines for trustworthy AI"~\cite{EUEthicsGuidelines} in 2019, and in 2021 proposed a draft for an AI Act, which sets out to regulate AI technology such as foundation models. If the current draft of EU AI Act, which passed through the European Parliament in June 2023, comes into effect some LLM's (e.g., ChatGPT) will not be in compliance with the regulation due in part to non-disclosure of the training data that is copyrighted~\cite{AIAct}. What then happens to the various services that organisations are building on top of the likes of ChatGPT?

\subsection{Transparency} 

In terms of the trustworthiness of foundation model AI technology, there are a number of deficits that we have already touched upon related to security and privacy, rights violations, harmful and factually incorrect content generation etc. Here we focus on transparency, or lack there of. This lack of transparency exists in a number of ways. Firstly, in terms of the lack of transparency around the data upon which many of these models are trained upon, secondly in terms of the associated practices (e.g., potentially exploitative labour practices) that might be used building guard rails and in training the RLHF. Thirdly, in terms of the lack of documentation and disclosures describing many of the commercial systems that are currently available. And fourthly, in terms of the inherent nature of the underlying foundation model technology of deep learning.

In the fourth case this exposes a fundamental limitation of deep learning technology, which has given rise to a significant open area of research (e.g., the sub-field of explainable AI). At one level we can describe the current generation of foundation model technology to be black-box in nature. That is, they receive some input (e.g., a user prompt) and generate a response (e.g., generated text), but we have no human-interpretable explanation of how the underlying model generated the response from the output. It is technically possible, and extremely laborious, to observe each of the millions or billions of low-level calculations that occur in generating the response, but these are so abstracted away from concepts in the real world that they are effectively meaningless and reveal no true explanatory insight. This raises serious questions and limitations in how the technology should be deployed, and touches again on the issue of regulatory non-compliance. If a foundation model is used as part of some decision-making system (whether automated or not) this means it will be extremely challenging, if not impossible, to extract an explanation for each decision or recommendation made. By way of example, if we are to consider existing regulations such as the EU GDPR~\cite{GDPR}, European citizens have a right to explanation of automated decisions that significantly affects them or produces legal effects, and they have the right to challenge those decisions. If a foundation model is part of such a decision-making process how can a citizen be provided with a meaningful explanation? 


\subsection{Equality, Diversity \& Inclusion}
Finally, we also need to be aware that the field of AI is (or at least, should be) inherently multidisciplinary. It draws upon a number of foundational disciplines including computer science, robotics, neuroscience, statistics and mathematics, psychology, linguistics, philosophy, biology, evolution and genetics, complex systems, humanities and social sciences, ethics and law, to name a few. At its best AI is an inherently interdisciplinary science, bringing researchers from all of these communities together in a collaborative effort. To truly understand intelligence, to build an artificial intelligence equivalent to (or even superior to) that of living organisms, and to understand the limitations and implications of the current state of the technology arising from the field of AI we need to draw upon the expertise of researchers from these core disciplines. In terms of large language models in particular, we would suggest that at a minimum we should listen to experts in linguistics (e.g., Prof Emily Bender and collaborators), statistics and causal inference (e.g., Prof Judea Pearl), and of course machine learning generally. 

This points to another potential limitation of many existing LLMs. They and their associated tools have not always been developed, tested and deployed by diverse teams. Diversity in this case meaning inclusion of the diverse set of core disciplines of expertise that should make up AI, in addition to the other desirable diversity attributes of a team designed to better represent a global society, including ethnicity, race, gender, sexual orientation, age, religious beliefs, socio-economic status, civil status, physical impairment, learning difficulties, and neurodiversity. We might argue, that if the teams in organisations responsible for developing and deploying AI technology such as LLMs were truly diverse, that many of the limitations and resulting issues of the technology might be more successfully addressed.

\section{Conclusions}

Given the level of hype surrounding Large Language Models and ChatGPT in particular, it is important to maintain a sense of balance and perspective. 
We have to zoom out and think of the bigger picture here. If a single company invests over 11 Billion dollars in a technology, what are the expectations of that company? For context, the Irish Governments Budget package for 2023 was 11 Billion Euro. In other words, you could run a small country with that kind of money. Arguably, with that level of investment, an organisation will want to maximise their return. Naturally, any signs of utility are going to be exploited and marketed. For example, perhaps they might allude to the magical properties of the technology, suggesting that everyone needs to get their hands on it to succeed in business and life more generally. The fact that the technology being hyped is itself capable of generating stories, blogs, and tweets at the speed of light is probably helping to feed the hype. They might allude to the potential of the technology to lead to what are science fiction Hollywood scenarios of AI being a threat to the existence of humanity and taking over the universe. If these organisations truly believe this to be a potential reality, then (at this moment in time) they have the power to stop it from happening. As they alone have the computing and data resources available at sufficient scale to realise an AI technology at the level of a ChatGPT, never mind at the level of what might be an artificial super intelligence. They could choose not to develop the technology, rather than call for others to stop them through regulation. However, whether we like it or not, the genie has left the bottle, and the technology is out in the wild. As such, we need to adapt to live with it and equip ourselves with some understanding of both the opportunities and the limitations of this technology as it evolves.

Ironically, LLM technology also has the potential to pollute the future training data sets of the next generation of LLMs, and even the re-training of existing model architectures with even more garbage. This might result in a negatively reinforcing feedback loop that further propagates the generation of increasing amounts of misinformation, bias, hate speech etc. As the machine learning adage goes, "garbage in, garbage out". In other words, if you train your model on data that is inappropriate for the task at hand, the resulting behaviour or output of the model will be useless.

\begin{figure}
    \centering
    \caption{ChatGPT's advice on when it should not be used}
    \label{chatgpt_use_advice}
    \includegraphics[angle=270,width=1.0\textwidth]{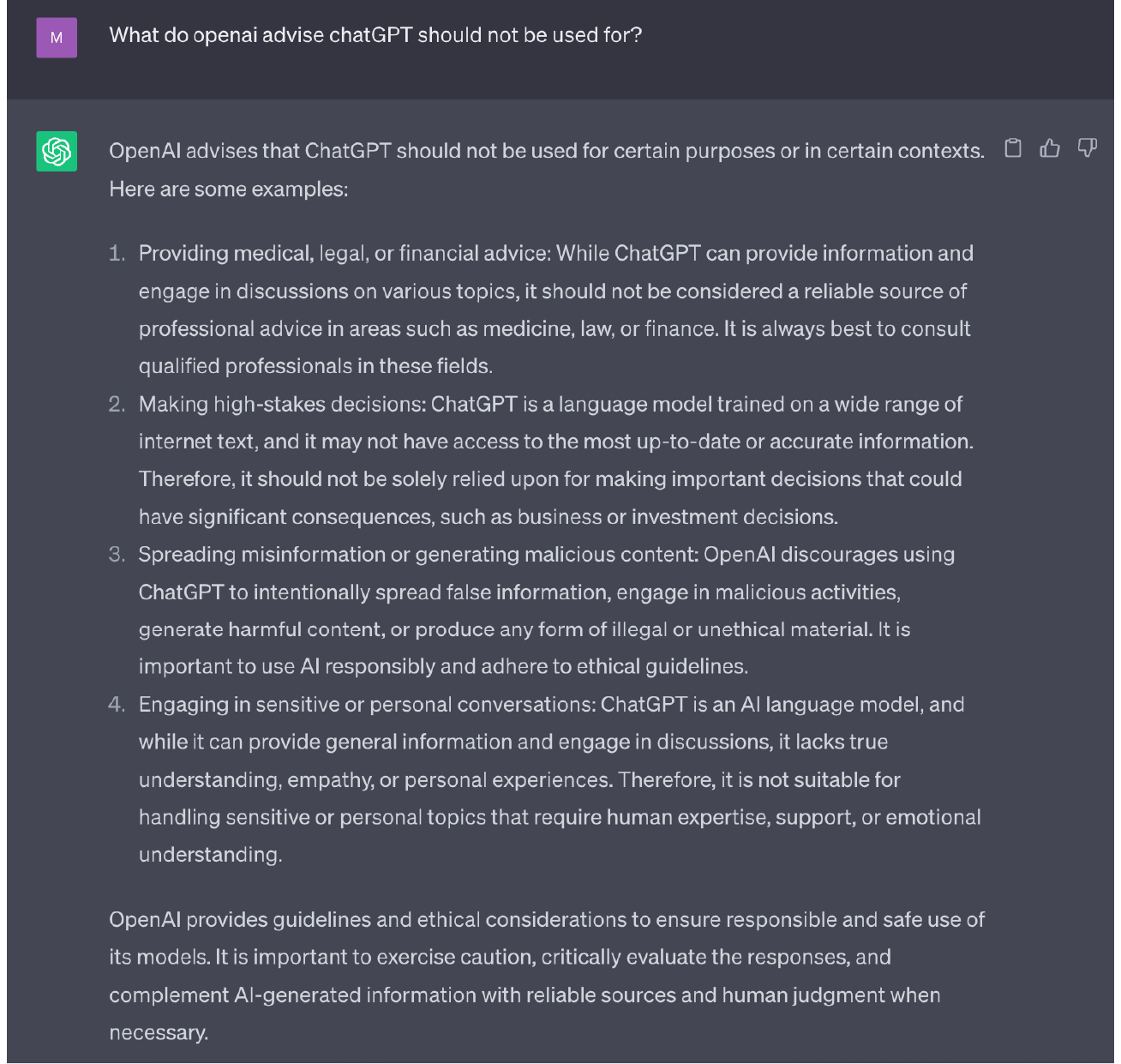}
\end{figure}

We have to constantly remind ourselves, that for the current generation of foundation model technology, no company that sells or uses this technology can guarantee that all the output generated will be factually correct, that it won't contain bias and/or harmful content WITHOUT having a HUMAN read it. Everyone NEEDS to read the small print. The creators of these models basically admit this to be the case, for example, see OpenAI's Usage Policy~\cite{OpenAIUsagePolicy}, which, does not permit ChatGPT use for legal and financial advice, or in health for diagnosis, treatment advice or disease management, or for government decision-making. We asked ChatGPT itself what it should not be used for, and the generated response is provided in Figure~\ref{chatgpt_use_advice}. In other words, we need a working language model (e.g., a human) to check the output of the AI language model. It will take an extraordinary effort and perhaps an infinite amount of time for, for example, reinforcement learning with human feedback to provide the necessary "guard rails" with an equivalence to a humans language model, or for humans to manually codify all the rules required! In the meantime, we need humans to be in the loop and always cognisant of the limitations, harms and risks of large language models. 







\section*{Acknowledgements}
The authors would like to thank Anthony Brabazon and Miguel Nicolau for invaluable discussions that have helped to inspire the drafting of this paper.

\end{document}